\documentclass[sigconf]{acmart}
\usepackage{color}
\usepackage{balance}
\usepackage{soul}
\usepackage{enumitem}
\usepackage{booktabs}
\usepackage{multirow}
\usepackage{siunitx}
\usepackage{fontawesome}
\usepackage[font=small,format=plain, justification=justified,singlelinecheck=false]{caption}

\begin{document}



\setcopyright{acmlicensed}
\acmConference[Neu-IR '17]{Neu-IR 2017 SIGIR Workshop on Neural Information Retrieval}{August 11, 2017}{Shinjuku, Tokyo, Japan} 
\acmYear{2017}
\copyrightyear{2017}
\acmDOI{}
\acmISBN{}
\acmPrice{}

\title{A Sentiment-and-Semantics-Based Approach for Emotion Detection in Textual Conversations}


\author{Umang Gupta}
\affiliation{%
  \institution{Microsoft, Hyderabad, India}
}
\email{umangup@microsoft.com}

\author{Ankush Chatterjee}
\authornote{Work done during Research Internship at Microsoft, India}
\affiliation{%
  \institution{IIT Kgp, Kharagpur, India}
}
\email{ankushchatterjee@iitkgp.ac.in}

\author{Radhakrishnan Srikanth}
\affiliation{%
  \institution{Microsoft, Hyderabad, India}
}
\email{rsrikan@microsoft.com}

\author{Puneet Agrawal}
\affiliation{%
  \institution{Microsoft, Hyderabad, India}
}
\email{punagr@microsoft.com}

\begin{abstract}
Emotions are physiological states generated in humans in reaction to internal or external events. They are complex and studied across numerous fields including computer science. As humans, on reading ``Why don't you ever text me!'' we can either interpret it as a sad or angry emotion and the same ambiguity exists for machines. Lack of facial expressions and voice modulations make detecting emotions from text a challenging problem. However, as humans increasingly communicate using text messaging applications, and digital agents gain popularity in our society, it is essential that these digital agents are emotion aware, and respond accordingly.

In this paper, we propose a novel approach to detect emotions like happy, sad or angry in textual conversations using an LSTM based Deep Learning model. Our approach consists of semi-automated techniques to gather training data for our model. We exploit advantages of semantic and sentiment based embeddings and propose a solution combining both. Our work is evaluated on real world conversations and significantly outperforms traditional Machine Learning baselines as well as other off-the-shelf Deep Learning models.

\end{abstract}

\fancyhead{}

\maketitle




\section{Introduction}
\label{sec:intro}
\begin{figure}[!t]
\centering
\includegraphics[scale = 1.0]{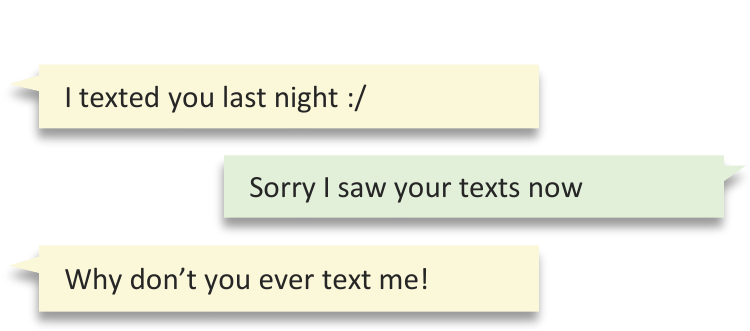}
\vspace{-0.1cm}
\caption{\small A sample 3-turn conversation from our dataset.}
\vspace{-0.4cm}
\label{chat}
\end{figure} 
Emotions are basic human traits and have been studied by researchers in the fields of psychology, sociology, medicine, computer science etc. for the past several years. Some of the prominent work in understanding and categorizing emotions include Ekman's six class categorization \cite{ekman1992argument} and Plutchik's ``Wheel of Emotion'' which suggested eight primary bipolar emotions \cite{plutchik1986emotion}. Given the vast nature of study in this field, there is naturally no consensus on the granularity of emotion classes. In this paper we consider 3 emotion classes, Happy, Sad, Angry along with an Others category; and classify a textual conversation into one of the above four. \\

\textbf{Problem Definition:} \emph{Given a textual user utterance along with 2 turns of context in a conversation, classify the emotion of user utterance as Happy, Sad, Angry or Others.}\\ 

Detecting emotions in textual conversations can be a challenging problem in absence of facial expressions and voice modulations. Figure \ref{chat} provides an example where it is difficult, even as a human, to detect the emotion of user utterance solely on the basis of text of the conversation. The emotion of the user whose messages are on the left, could be interpreted as angry or sad. The challenge of detecting emotions is further compounded by difficulty in understanding context, sarcasm, class size imbalance, natural language ambiguity and rapidly growing Internet slang.

However, with the growing prominence of messaging platforms like WhatsApp and Twitter as well as digital agents, it is essential that machines are able to understand emotions in textual conversations and avoid responding inappropriately \cite{miner2016smartphone}. Emotion detection technology can find several applications in today's online world. In domain of customer service, social media platforms like Twitter are gaining prominence where customers expect quick responses. In case of heavy flow of tweets, turn-around time for responses increase. If tweets can be prioritized according to their emotional content and responded to in that order, it will increase customer satisfaction. For example, responding to an angry tweet prior to a basic inquiry. Also, in this era of text messaging, users are constantly texting and may send inappropriately angry messages to others. If emotion detection is implemented, in such cases, the application can take appropriate action such as popping up a warning to the user before sending a message.

In this paper, we propose a deep learning approach for detecting emotions in textual conversations. We use sentiment and semantic representation of text to create a unified LSTM architecture called ``Sentiment and Semantic LSTM (SS-LSTM)'' to detect emotions.
Our model SS-LSTM does not require hand-crafted features and is trained as a single unified model. We evaluate SS-LSTM on real world textual conversations and it outperforms traditional Machine Learning approaches and other Deep Learning based approaches. The main contributions of our paper are as follows:
\begin{itemize}[noitemsep]
\item We propose a novel deep learning approach called ``Sentiment and Semantic LSTM (SS-LSTM)'' to detect emotions in textual conversations. 
\item We evaluate various Deep Learning techniques and embeddings, along with Machine learning algorithms (such as SVM, Decision Trees, Naive Bayes), on real world textual conversations and compare their effectiveness for the task of detecting emotions.
\end{itemize}

The rest of the paper is organized as follows: Section 2 provides a summary of related work. Section 3 describes our approach (SS-LSTM) in detail. Our experimental setup is discussed in Section 4 and our results are in Section 5. Section 6 concludes the paper and finally Section 7 has acknowledgements.

\section{Related Work} 
\label{sec:related}
\begin{figure}[!t]
\centering
\includegraphics[scale = 0.6]{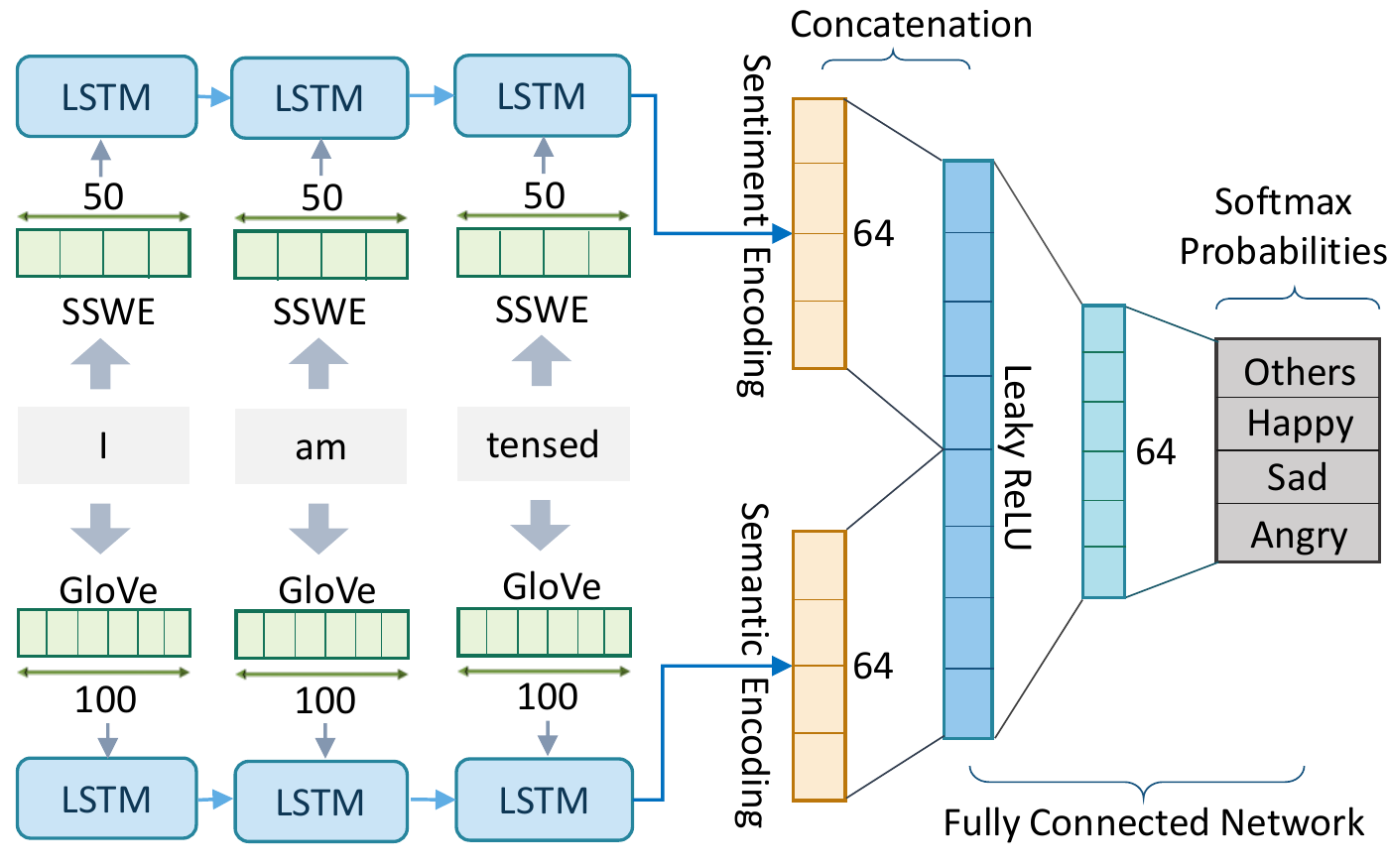}
\vspace{-0.1cm}
\caption{\small The architecture of Sentiment and Semantic LSTM (SS-LSTM) Model.}
\vspace{-0.2cm}
\label{architecture}
\end{figure} 

The field of sentiment analysis has been extensively studied. However, limited research exists in classifying textual conversations based on emotions. In a bid to gather annotated data on a large scale, some researchers have used automated methods such as emoticons, sentiment analysis and hashtags to label the data \cite{hasan2014hashtag, hasan2014emotex, purver2012experimenting, suttles2013distant, wang2012harnessing}. Our method relies on a combination of Deep Learning based data expansion, heuristics and human judgment to create a large corpus of training data for the model. 

Emotion-detection algorithms can broadly be categorized into 3 classes: \\
(a) \emph{Rule based approaches} - Some methods  exploit the usage of keywords in a sentence and their co-occurrence with other keywords with explicit emotional/affective value \cite{kozareva2007ua,balahur2011detecting, chaumartin2007upar7, strapparava2008learning, sykora2013emotive}. To that effect, several lexical resources are used, some of the most popular ones being WordNet-Affect \cite{strapparava2004wordnet} and SentiWordNet \cite{esuli2007sentiwordnet}. Part-of-Speech taggers like the Stanford Parser are also used to exploit the structure of keywords in a sentence. Such methods often need hand-crafting and have good precision, but suffer from low recall since many sentences do not contain affective words despite conveying emotions, e.g. ``Trust me! I am never gonna order again''.
Our method differs from such Rule based approaches as it does not require any hand-crafted features, which are often unable to capture all possible representations of emotions. \\
(b)	\emph{Non-neural Machine Learning approaches} -  For sentiment analysis as well as emotion detection, most methods rely on extracting features such as presence of frequent n-grams, negation, punctuation, emoticons, hashtags etc. to form a feature representation of the sentence, which is then used as input by classifiers such as Decision Trees, Naive Bayes, SVMs among others to predict the output \cite{alm2005emotions, balabantaray2012multi, davidov2010enhanced, kunneman2014predictability, suttles2013distant, yan2016exploring}. More detailed analysis have been provided in \cite{canales2014survey}. Vosoughi et al. \cite{vosoughi2015enhanced} extract tweets based on location, time and author and uses context to model prior in Bayesian models. These methods often require extensive feature engineering and do not achieve high recall due to diverse ways of representing emotions.\\
(c) \emph{Deep Learning approaches} - Deep Neural networks have enjoyed considerable success in varied tasks in text, speech and image domains. Variations of Recurrent Neural Networks, such as LSTM \cite{hochreiter1997long} and BiLSTM \cite{schuster1997bidirectional} have been effective in modeling sequential information. Also, Convolutional Neural Networks \cite{krizhevsky2012imagenet} have been a popular choice in the image domain. The lower layers of the network capture local features whereas higher layers unravel more abstract task based features for the image. Their introduction to the text domain has proven their ability to decipher abstract concepts from raw signals \cite{kim2014convolutional,prakash2016emulating}. One of the approaches employs CNNs to classify emotion features \cite{mundra2017fine}. The vast success of Deep Neural Nets and their ability to perform tasks without hand-crafting features is our motivation to try these techniques for detecting emotions. We combine both sentiment and semantic features from user utterance to improve emotion detection.

\section{Our Approach}
\label{sec:approach}
We model the task of detecting emotions as a multi-class classification problem where given a user utterance, the model outputs probabilities of it belonging to four output classes - Happy, Sad, Angry and Others. The architecture of our proposed SS-LSTM model is shown in Figure \ref{architecture}. The input user utterance is fed into two LSTM layers using two different word embedding matrices. One layer uses a semantic word embedding, whereas the other layer uses a sentiment word embedding.  These two layers learn semantic and sentiment feature representation and encode sequential patterns in the user utterance. These two feature representations are then concatenated and passed to a fully connected network with one hidden layer which models interactions between these features and outputs probabilities per emotion class. Further details of training data used to train the model, sentiment and semantic embeddings, and model training are provided below. 

\subsection{Training Data Collection}

\begin{table}[!t]
\centering
\small
  \begin{tabular}{p{0.8cm}ccccc}
  \toprule
	\textbf{Label} & \textbf{Happy} & \textbf{Sad} & \textbf{Angry} & \textbf{Others} & \emph{\textbf{Total}}\\
  \midrule
	 \textbf{\#} & 109 & 107 & 90 & 1920 & 2226\\
	 \textbf{\%} & 4.90 & 4.81 & 4.04 & 86.25 & 100\\
  \bottomrule
  \end{tabular}
\vspace{0.2cm}
\caption{\small Emotion class label distribution in evaluation dataset.}
\vspace{-0.2cm}
\label{data-stats}
\end{table}

Given the potentially diverse representation of emotions, we collected a large amount of training data using a semi-automated approach. We constructed a dataset of 17.62 million tweet conversational pairs i.e. tweets (Twitter-Qs) and their responses (Twitter-As; collectively referred to as Twitter Q-A pairs below), extracted from the Twitter Firehose, covering the four year period from 2012 through 2015. This data was further cleaned to remove twitter handles and served as the base data for our two training data collection techniques. \\
\\
\textbf{Technique 1: } 
In this technique, we start with a small set (approximately 300) of annotated utterances per emotion class obtained by showing a randomly selected sample from Twitter-Qs and Twitter-As to human judges. Using a variation of the model described in \cite{palangi2016deep}, we created sentence embeddings for these annotated utterances as well as Twitter-Qs and Twitter-As. We identified potential candidate utterances for each emotion class using the threshold-based cosine similarity between annotated utterances and Twitter-Qs and Twitter-As. Various heuristics like presence of opposite emoticons (example ``:'('' in a potential candidate set for Happy emotion class), length of utterances etc. are used to further prune the candidate set. The candidate set is then shown to human judges to determine whether or not they belong to the emotion class. Using this method we cut down the amount of human judgments required by five times when compared to showing a random sample of utterances and choosing emotion class utterances from them. \\
\\
\textbf{Technique 2: }
Once we obtain utterances belonging to an emotion class by the method described above, we take all the utterances that belonged to Twitter-Qs and find their corresponding Twitter-As. These Twitter-As are then further aggregated by their frequency and top Twitter-As are chosen. For example in the Angry emotion class ``There, there''\footnote{\small A phrase frequently used in popular American sitcom, ``The Big Bang Theory''} was a popular response in Twitter-As. Twitter-Qs corresponding to these top Twitter-As per emotion class are picked as potential utterances in that class and are further shown to human judges for pruning.\\

Negative data (belonging to class Others) is collected by randomly selecting utterances from both Twitter-Qs and Twitter-As. Those which have a high cosine score (using Technique 1) with any of the utterances in emotion classes (Happy, Sad, Angry) are discarded.

We finally obtained 456k utterances in the Others category, 28k for Happy, 34k for Sad, and 36k for Angry. 

\subsection{Emoticon Handling and Normalization}

Emoticons are frequently used in textual conversations. In Twitter Q-A pairs we found 21\% of textual conversations contain emoticons. We used several heuristics and normalization techniques to specifically deal with emoticons. For example, we converted the following utterance ``Yeah! :((( My plan is cancelled \faMehO\faFrownO'' into ``Yeah! :( My plan is cancelled :| :(''. This helps us deal with Out of Vocabulary (OOV) issues for infinitely many possible combinations of emoticons, and convert various forms of emoticons which represent similar feelings to a singular form.

\subsection{Choosing Input Embeddings}

\begin{table}[t]
\centering
  \begin{tabular}{lcccc}
    \toprule
    \multirow{2}{*}{} &
      \multicolumn{1}{c}{\textbf{\textsc{Happy}}} &
      \multicolumn{1}{c}{\textbf{\textsc{Sad}}} &
      \multicolumn{1}{c}{\textbf{\textsc{Angry}}} &
       \\
      
       & {F1} & {F1} & {F1} & {Avg. F1} \\
      \midrule 
	  Word2Vec & 64.44 & 74.71 & 59.28 & 66.14 \\
	  FastText & 64.58 & 76.68 & 59.98 & 67.08 \\
      GloVe	& 66.11 & 78.99 & 63.79 & 69.63 \\
      SSWE & 65.64 & 78.22 & 63.22 & 69.32 \\

    \bottomrule
  \end{tabular}
\vspace{0.2cm}
\caption{\small Comparison of results obtained from different embeddings using an LSTM network.}
\vspace{-0.5cm}
\label{embedding-results}
\end{table}

\begin{table}[t]
\centering
  \begin{tabular}{p{3cm}rr}
    \toprule
	\textbf{Word1, Word2} & \textbf{GloVe} & \textbf{SSWE} \\     
      \midrule
	  depression, :'{(} & 0.23 & 0.63 \\
      happy, sad & 0.59 & -0.42 \\
      best, great & 0.78 & 0.15 \\

    \bottomrule
  \end{tabular}
\vspace{0.2cm}
\caption{\small Comparison of GloVe and SSWE embeddings w.r.t cosine similarity of word pairs.}
\vspace{-0.2cm}
\label{sswe-glove-cosine}
\end{table}

For each word in the input utterance we obtain word embeddings using several techniques. We try Word2Vec \cite{mikolov2013w2v}, GloVe \cite{pennington2014glove}, FastText \cite{joulin2016fasttext} as well as Sentiment Specific Word Embedding (SSWE) \cite{tang2014sswe}. SSWE aims at encoding sentiment information in the continuous representation of words. To test the effectiveness of these embeddings for emotion detection, we train a simple Long-Short Term Memory (LSTM) model using each of these embeddings. LSTMs are variants of Recurrent Neural Networks (RNN) \cite{hochreiter1997long} and have the ability to capture long-term dependencies present in the input sequence, and thus are helpful for our task. We use cross validation to determine the effectiveness of different embeddings. Our results, as depicted in Table \ref{embedding-results}, indicate that GloVe gives the best average F1 score which is slightly better than SSWE F1 score. However, we also observe that GloVe and SSWE behave very differently; a few examples can be seen in Table \ref{sswe-glove-cosine}. SSWE embeddings give a high cosine similarity when calculated for ``depression'' and ``:'(''  whereas GloVe gives a low score even though the two words have similar sentiment. For the ``happy'' and ``sad'' pair, SSWE rightly gives a low score but GloVe outputs a reasonably high score. However, semantically similar words like ``best'' and ``great'' have a low cosine similarity with SSWE but high score from GloVe. Based on these observations, we choose GloVe as our embedding for the Semantic LSTM layer and SSWE as our embedding for the Sentiment LSTM layer.

\begin{table*}[!t]
\centering
\resizebox{\textwidth}{!}{%
  \begin{tabular}{lrrrrrrrrrr}
    \toprule
    \multirow{2}{*}{} &
      \multicolumn{3}{c}{\textsc{\textbf{Happy}}} &
      \multicolumn{3}{c}{\textsc{\textbf{Sad}}} &
      \multicolumn{3}{c}{\textsc{\textbf{Angry}}} &
      \\
      & \textsc{Precision} & \textsc{Recall} & \textsc{F1} & \textsc{Precision} & \textsc{Recall} & \textsc{F1} & \textsc{Precision} & \textsc{Recall} & \textsc{F1} & \textsc{Avg. F1} \\
      \midrule
	  NB & 41.35 & 50.46 & 45.45 & 70.87 & 68.22 & 69.52 & 38.16 & 32.22 & 34.94 & 49.97 \\
      SVM & 66.67 & 25.69 & 37.09 & 86.49 & 59.81 & 70.71 & 85.42 & 45.56 & 59.42 & 55.74 \\
      GBDT & 75.76 & 22.94 & 35.21 & 89.47 & 63.55 & 74.31 & 86 & 47.78 & 61.43 & 56.98 \\
      \midrule
      CNN-NAVA & 63.32 & 42.29 & 50.71 & 79.37 & 68.69 & 73.64  & 67.42 & 45.79 & 54.54 & 59.63  \\ 
      \midrule
      CNN-SSWE & 67.69 & 40.37 & 50.57 & 77.45 & 73.83 & 75.6 & 80.95 & 37.77 & 51.51 &	59.23 \\
      CNN-GloVe & 52.29 & 52.29 & 52.29 & 93.72 & 67.29 & 74.61 & 67.82 & 65.55 & 66.66 & 64.52 \\
      LSTM-SSWE & 70.69 & 37.61 & 49.1 & 83.87 & 72.89 & 78 & 73.24 & 57.77 & 64.6 & 63.9 \\
      LSTM-GloVe & 64.18 & 39.45 & 48.86 & 72.88 & 80.37 & 76.44 & 72.15 & 63.33 & 67.45 & 64.25 \\
      \midrule
      SS-LSTM & 69.51 & 52.29 & \textbf{59.68} & 85.42 & 76.63 & \textbf{80.79} & 87.69 & 63.33 & \textbf{73.55} & \textbf{71.34} \\    
    \bottomrule
  \end{tabular}}
\vspace{0.2cm}
\caption{\small Comparison of various models on evaluation dataset. SS-LSTM results are statistically significant with p < 0.005}
\vspace{-0.2cm}
\label{model-results}
\end{table*}

\subsection{Model Training}
We use the Microsoft Cognitive Toolkit\footnote{\small https://www.microsoft.com/en-us/cognitive-toolkit/} for training SS-LSTM. The parameters of SS-LSTM are trained to maximize prediction accuracy given the target labels in the training set. We split our training data in a 9:1 ratio to create sets for training and validation respectively. We train the model using the training set and tune the hyper-parameters using the validation set. We use Cross Entropy with Softmax as our loss function \cite{goodfellow2016deep}, and Stochastic Gradient Descent (SGD) as our learner. We found the optimal batch size to be 4000 with a learning rate of 0.005. It is worth noting that when training sequence classification models, the Microsoft Cognitive Toolkit uses the sum of the length of sequences across utterances (not the number of utterances) when picking up data of a particular batch size. 

\section{Experimental Setup}
\label{sec:experiment}
In this section we describe details of evaluation dataset used to compare various techniques and baseline methods used for comparison.  

\subsection{Evaluation Dataset}
We are aware of two datasets in this domain: (a) The ISEAR dataset\footnote{\small http://www.affective-sciences.org/en/home/research/materials-and-online-research/research-material/} and (b) The SemEval2007 Affective Text Dataset\footnote{\small http://nlp.cs.swarthmore.edu/semeval/tasks/task14/data.shtml}. However, both these datasets are unsuitable for evaluating our task. ISEAR dataset consists of user reactions when they were asked to remember a circumstance which aroused certain emotions in them. For example ``When my mother slapped me in the face, I felt anger at that moment.'' is one of the statements in ISEAR dataset and has a different form than what one would expect in a conversation. On the other hand, SemEval2007 dataset consists of news headlines which are again not similar to conversations. 

To overcome these challenges we sample 3-turn conversations from Twitter i.e. User 1's tweet; User 2's response to the tweet, and User 1's response to User 2. We used the Twitter Firehose to extract these 3 turn conversations covering the year of 2016. We sampled from conversations where the last turn was the third turn as well as from those where the third turn was in the middle of the conversation. Our dataset finally comprised of 2226 3-turn conversations along with their emotion class labels (Happy, Sad, Angry, Others) provided by human judges. The details of the dataset along with emotion class label statistics is shown in Table \ref{data-stats}. To gather the emotion class labels, we showed the third turn of the conversation along with the context of the previous 2 turns to human judges and asked them to mark the emotion of the third turn after considering the context. To gather high quality judgments each conversation was shown to 5 judges, and a majority vote was taken to decide the emotion class. After several rounds of training and auditing of mock sets, the final inter-annotator agreement based on fleiss' kappa value \cite{shrout1979kappa} was found to be 0.59. This kappa value, while slightly less then desirable, indicates the difficulty in judging textual conversations due to ambiguities discussed earlier in Section 1. 

Our evaluation dataset is unseen at time of training. SS-LSTM and all baseline approaches are evaluated on this dataset.

\begin{table*}[!t]
\centering
\resizebox{\textwidth}{!}{%
  \begin{tabular}{llp{3cm}p{3.5cm}p{3.7cm}p{6cm}}
    \toprule
    \textbf{\#} & \textbf{True Label}  & \textbf{User 1's tweet}  
    & \textbf{User 2's response} & \textbf{User 1's response} & \textbf{Comment} \\
    \midrule
      1
      & Angry
      & It will be arranged within two business days?
      & It will be done at the earliest
      & This is getting very annoying now, no pickup yet!
      & LSTM-SSWE and SS-LSTM predict correctly, probably because of the keyword `annoying' which represents negative sentiment. LSTM-GloVe fails.  \\
      
      \midrule
      
      2
      & Sad
      & Man even food delivery apps in bangalore won't deliver till 6:(
      & Yea well it is a bandh
      & Yeah well i do not have anything at home :/
      & LSTM-SSWE fails as there is no keyword with an obvious negative polarity but LSTM-GloVe and SS-LSTM are correct.  \\
      
      \midrule
      
      3
      & Angry 
      & :{)} Good for both of us!
      & It's better not to interact with a girl with so much ego. Attitude is still fine
      & It is not an ego or attitude. U started first! U asked me stupid ques! :/
      & SS-LSTM is only model which could correctly predict this rather complicated user utterance.\\
      
      \midrule
      
      4
      & Sad
      & 3 gone 2 more to go :3
      & Crores? :D
      & Haha no ya. My kittens. One by one they are all leaving :'(
      & Presence of `Haha' and ``:'('' make this case difficult to predict and all models including SS-LSTM fail \\
      
      \midrule
      
      5
      & Happy
      & I just qualified for the Nabard internship
      & WOOT! That's great news. Congratulations!
      & I started crying 
      & All models predicted it as Sad, however, when one takes into account context, true emotion is Happy.\\
      
    \bottomrule
  \end{tabular}}
\vspace{0.2cm}
\caption{\small Qualitative Analysis of SS-LSTM results and other baseline approaches.}
\vspace{-0.24cm}
\label{qualitative-analysis}
\end{table*}

\subsection{Baseline Approaches}
We compare our approach against two classes of baselines. (a) Machine Learning based baselines and (b) Deep Learning based baselines.

For Machine Learning based baselines we used a Support Vector Machine (SVM) classifier \cite{cortes1995support}, a Gradient Boosted Decision Tree (GBDT) classifier \cite{friedman2001elements} and a Naive Bayes (NB) classifier \cite{friedman2001elements}. SVM, GBDT and NB classifiers were trained using Scikit Learn \cite{pedregosa2011scikit}. One of the salient features of our approach is the lack of need for feature engineering. Hence we kept the feature set small for SVM, GBDT and NB. We used 1,2,3 n-grams as features along with a hand-crafted emoticon feature set. This emoticon feature set is a 3 dimensional vector where the first dimension is the count of Happy emoticons like ``:)'' in the utterance. Similarly the second and third dimension are for Sad and Angry Emotions. After tuning parameters using the validation set as described in Section 3.4 we found SVM to give the best performance with linear Kernel and regularization constant 0.005. In case of GBDT, the best performance was achieved with 50 trees and a minimum of 10 samples per leaf.

For deep learning based baseline we implemented the approach defined in \cite{mundra2017fine}. To the best of our knowledge this work is the only other deep learning based approach attempted to detect emotion classes. We call this approach CNN-NAVA. We trained emotion vectors as defined in \cite{agrawal2012emotionvec} and used them as input to a CNN model. We also trained individual CNN and LSTM models with different embeddings like GloVe and SSWE. 

We used Precision, Recall, F1 score, and Average F1 score (where average is taken across F1 scores of emotion classes i.e. happy, sad and angry) to evaluate different approaches.

\section{Results}
\label{sec:results}

A summary of results from various techniques on the dataset described in Section 4.1 is presented in Table \ref{model-results}. SS-LSTM gives the best performance on F1 score for each emotion class as well as on Average F1. The performance of SS-LSTM over all other models is particularly significant (p < 0.005) as measured by McNemar's test \cite{mcnemar1969psychological}. Our results thus indicate that combining sentiment and semantic features in SS-LSTM outperforms individual LSTM-SSWE and LSTM-GloVe. SS-LSTM was also significantly better than CNN based approaches including CNN-NAVA. Also, when comparing across models using Average F1 score, Deep Learning based models outperform NB, SVM and GBDT. 

\subsection{Qualitative Analysis}
Table \ref{qualitative-analysis} highlights some examples from evaluation set and compares the performance of our models across these examples.
We observe that if user utterance had keywords or emoticons with a certain sentiment polarity associated with them, LSTM-SSWE usually works well even if LSTM-GloVe does not. The absence of the same affected LSTM-SSWE's performance. SS-LSTM, by combining both the feature sets, is able to accurately predict examples \#1-3 in Table \ref{qualitative-analysis}. Specifically, in \#3, all baseline approaches fail, but SS-LSTM is able to harness the advantage of combining both semantic and sentiment features to predict it correctly. However, SS-LSTM still needs further improvement. For example in \#4 presence of keywords like ``Haha'' and ``:'('' make it difficult for all models to predict it correctly. In some utterances like in \#5 context of the conversation plays an important role to determine underlying emotion, SS-LSTM does not consider context and hence fails as do all other models.

\begin{table}[!t]
\centering
\small
  \begin{tabular}{lp{2.3cm}p{2.2cm}p{2.1cm}}
    \toprule
    \textbf{\#} & \textbf{User 1}  
    & \textbf{User 2} & \textbf{User 1} \\
    \midrule
      1
      & Good morning! weekend
      & Good morning. :) :) :) :)
      & Happy Morning \\
      
      \midrule
      
      2
      & What r the birthday plans? ;{)}
      & going to hills with friends.
      & Oh great! \\
      \midrule
      3
      & I had a match today.
      & And did you win?
      & Yes!! And I am super happy :{)} \\

    \bottomrule
  \end{tabular}
\vspace{0.2cm}
\caption{\small Sample conversations indicating challenges in Happy emotion class}
\vspace{-0.4cm}
\label{happy-qualitative-analysis}
\end{table}

\subsection{Discussion on Ambiguity in Happy Class}
On comparing the F1 scores of several models in Table \ref{model-results} we observe that the Happy emotion class performs significantly worse than other emotion classes. We found inter-judge agreement to be particularly low for the Happy emotion class, which indicates variation in how a user utterance is interpreted by different human judges. In example \#1 of Table \ref{happy-qualitative-analysis}, User 1's second utterance is interpreted as Happy by some judges and just as a greeting by some other judges who mark it as Others. Similarly in example \#2, User 1's second utterance is considered a comment by some and happy statement by others due to the keyword ``great".  While in example \#3 User 1 is visibly happy, which is marked Happy by most judges. We thus believe that predicting utterances for the Happy class on basis of textual conversation alone is a challenging problem and hence, understanding context becomes even more important for this class.

\section{Conclusion}
\label{sec:conclusions}
We proposed a Deep Learning based approach called ``Sentiment and Semantic LSTM (SS-LSTM)'' to detect emotions in textual conversations. Our approach combines sentiment and semantic features from user utterance using SSWE and GloVe embeddings respectively and do not require any hand-crafted features. Evaluation on real world textual conversation shows that our approach outperforms CNN and LSTM baselines, in addition to other Machine Learning baselines. We observe that our approach can benefit from the ability to handle context. As part of future work, we plan to extend this approach to train models that are context aware.

\section{Acknowledgments}
We thank Balakrishnan Santhanam, Jaron Lochner and Rajesh Patel for their support in crowdsource judgments. We also thank Oussama Elachqar, Chris Brockett, Michel Galley, Manoj K Chinnkotla, Niranjan Nayak and Kedhar Nath Narahari for helpful brainstorming sessions and comments. Finally we are grateful to our peers Abhay Prakash and Meghana Joshi for their constant support and guidance.

\bibliographystyle{abbrv}
\bibliography{emoir}
\balance
\end{document}